\documentclass[10pt,twocolumn,letterpaper]{article}

\usepackage{3dv}
\usepackage{times}
\usepackage{epsfig}
\usepackage{graphicx}
\usepackage{amsmath}
\usepackage{amssymb}
\usepackage{multirow}

\newcommand{\norm}[1]{\left\lVert#1\right\rVert}
\newcommand{\boldparagraph}[1]{\vspace{0.2cm}\noindent{\bf #1:}}

\usepackage[pagebackref=true,breaklinks=true,letterpaper=true,colorlinks,bookmarks=false]{hyperref}

\threedvfinalcopy %

\def\threedvPaperID{93} %

\ifthreedvfinal\fi
\begin{document}

\title{Human Performance Capture from Monocular Video in the Wild}

\author{Chen Guo$^{1}$ \quad Xu Chen$^{1,2}$ \quad Jie Song$^{1}$ \quad Otmar Hilliges$^{1}$ \\
 $^1$ETH Z{\"u}rich \quad 
 $^2$Max Planck Institute for Intelligent Systems, T{\"u}bingen \\
}

\maketitle
\thispagestyle{empty}

\begin{abstract}
Capturing the dynamically deforming 3D shape of clothed human is essential for numerous applications, including VR/AR, autonomous driving, and human-computer interaction. Existing methods either require a highly specialized capturing setup, such as expensive multi-view imaging systems, or they lack robustness to challenging body poses. In this work, we propose a method capable of capturing the dynamic 3D human shape from a monocular video featuring challenging body poses, without any additional input. We first build a 3D template human model of the subject based on a learned regression model. We then track this template model's deformation under challenging body articulations based on 2D image observations. Our method outperforms state-of-the-art methods on an in-the-wild human video dataset 3DPW. Moreover, we demonstrate its efficacy in robustness and generalizability on videos from iPER datasets.
\end{abstract}

\section{Introduction}
\newcommand{\figureTeaser}{
\begin{figure}
\begin{center}
\includegraphics[width=\linewidth,trim=0 10 0 0, clip]{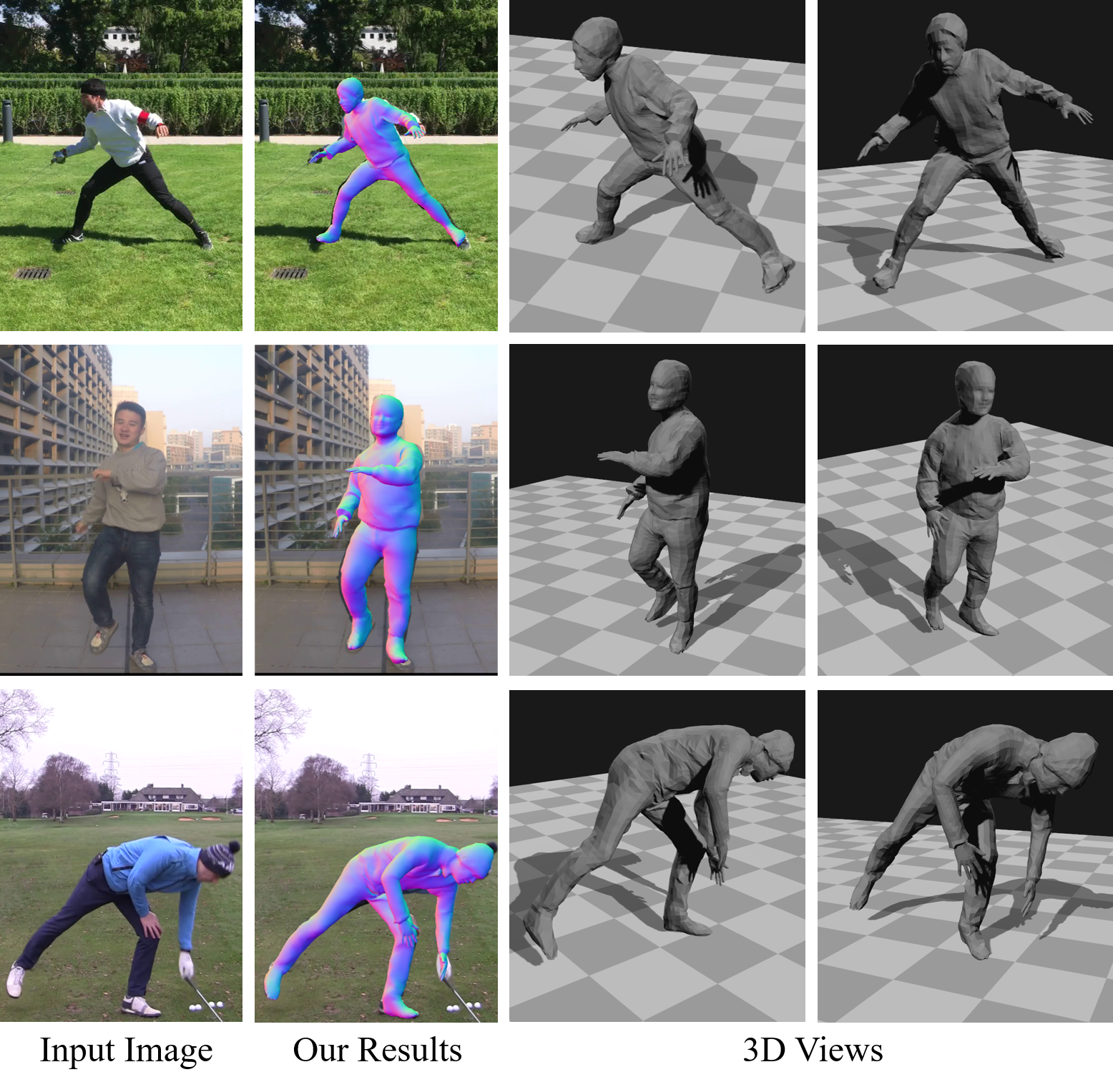}
\end{center}
\caption{\textbf{Human performance capture from monocular video in the wild}. We present a method to reconstruct the dynamically deforming 3D surface of human from a monocular video. Our method does not require a pre-scanned subject-specific template model and generalizes well to challenging poses, thus it is applicable in in-the-wild settings.}
\vspace{-2em}
\label{fig:teaser}
\end{figure}
}

\newcommand{\figurePipeline}{
\begin{figure*}
\begin{center}
\includegraphics[width=\linewidth]{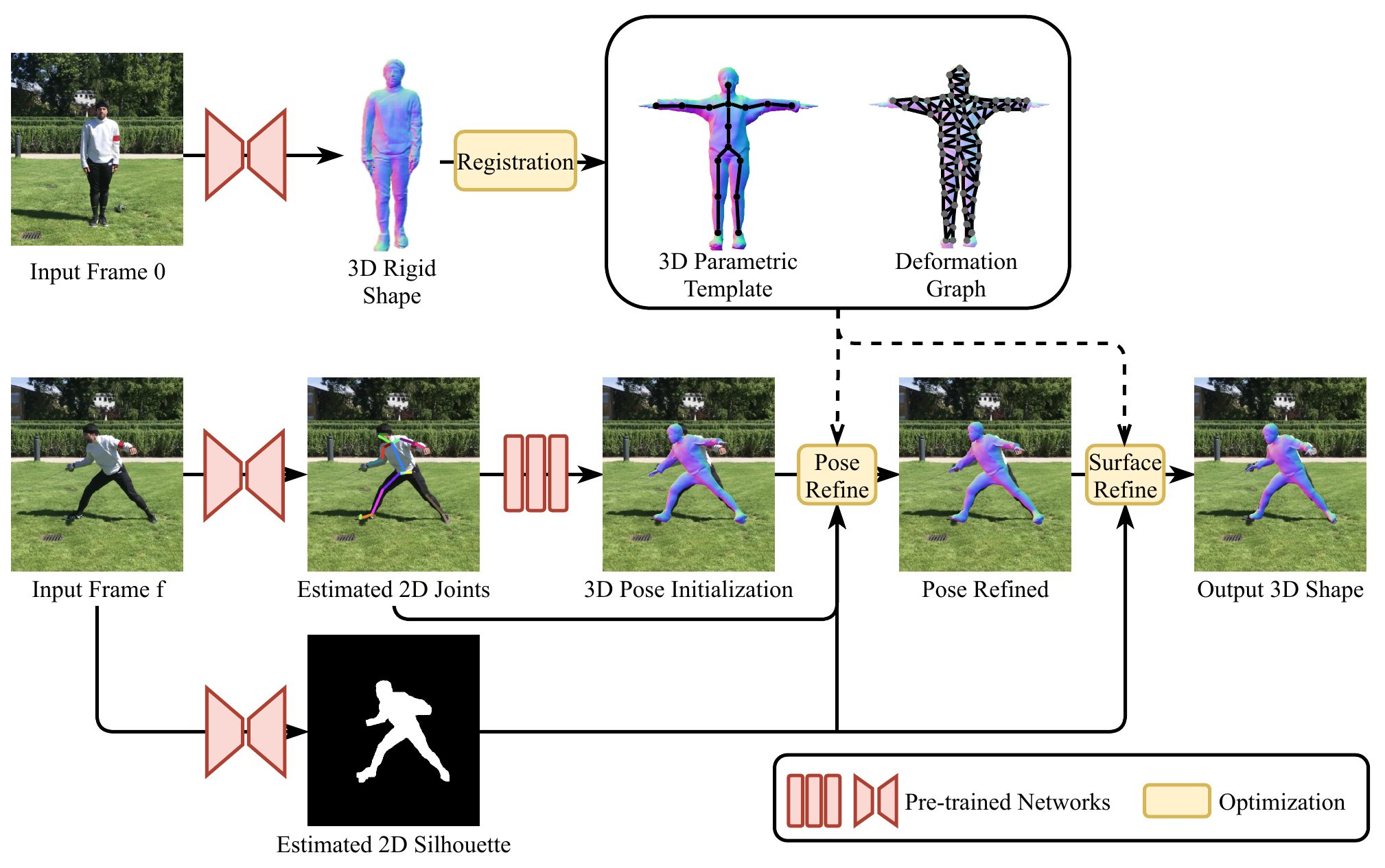}
\end{center}
\caption{\textbf{Method overview}. Given a monocular video as input, a parametric template model is automatically constructed from the initial frame of the video. We reconstruct the rigid 3D shape using the state-of-the-art single view reconstruction method \cite{saito2020pifuhd}. We then register a generic human model SMPL \cite{SMPL:2015} onto the rigid shape to build the parametric template with an embedded deformation graph. To reconstruct the dynamically deforming 3D surface at each frame, we optimize the pose, shape, and surface deformation parameters of the template to image observations. We first extract 2D joints and silhouettes from the RGB image. From 2D joints, we estimate 3D body poses using \cite{song2020lgd} to initialize the pose parameter. After initialization, we optimize the pose parameters by aligning the template with the 2D joints and silhouette. Afterward, we further optimize the detailed surface deformation by silhouette alignment.}
\label{fig:pipeline}
\end{figure*}
}

\newcommand{\figureAblationFinal}{

\begin{figure}
\begin{center}
\includegraphics[width=\linewidth,trim=0 10 0 0, clip]{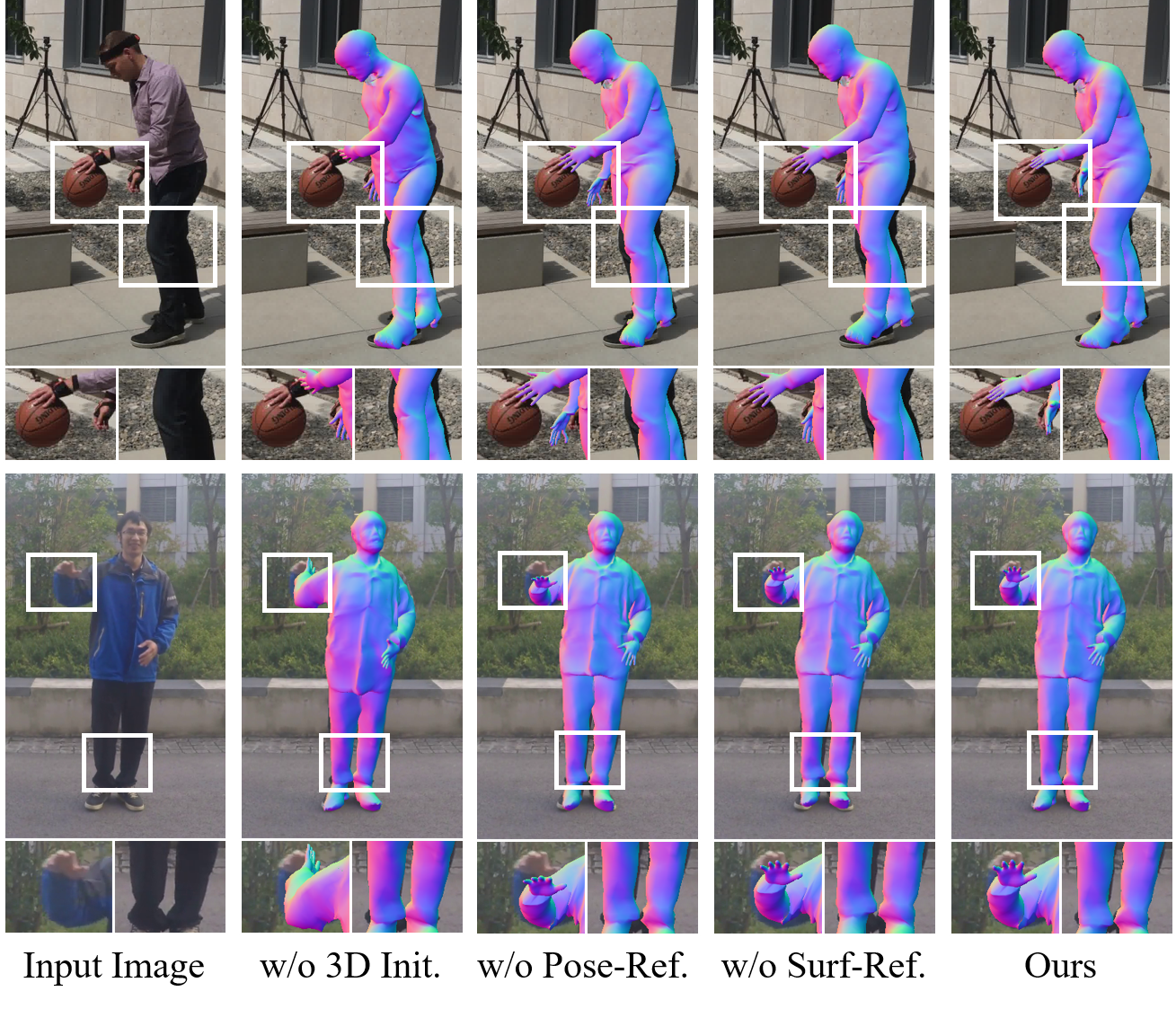}
\end{center}
\vspace{-1em}
\caption{\textbf{Qualitative evaluation of optimization stages.} Without a learned 3D pose initialization, the method tends to produce invalid poses due to accumulated error. Without pose refinement, the method suffers from noticeable misalignment to the image observation. Without surface refinement, the method fails to capture the non-linear deformations of the body and clothing.}
\label{fig:ablationfinal}
\end{figure}
}

\newcommand{\figureAblationStep}{
\begin{figure}
\begin{center}
\includegraphics[width=\linewidth]{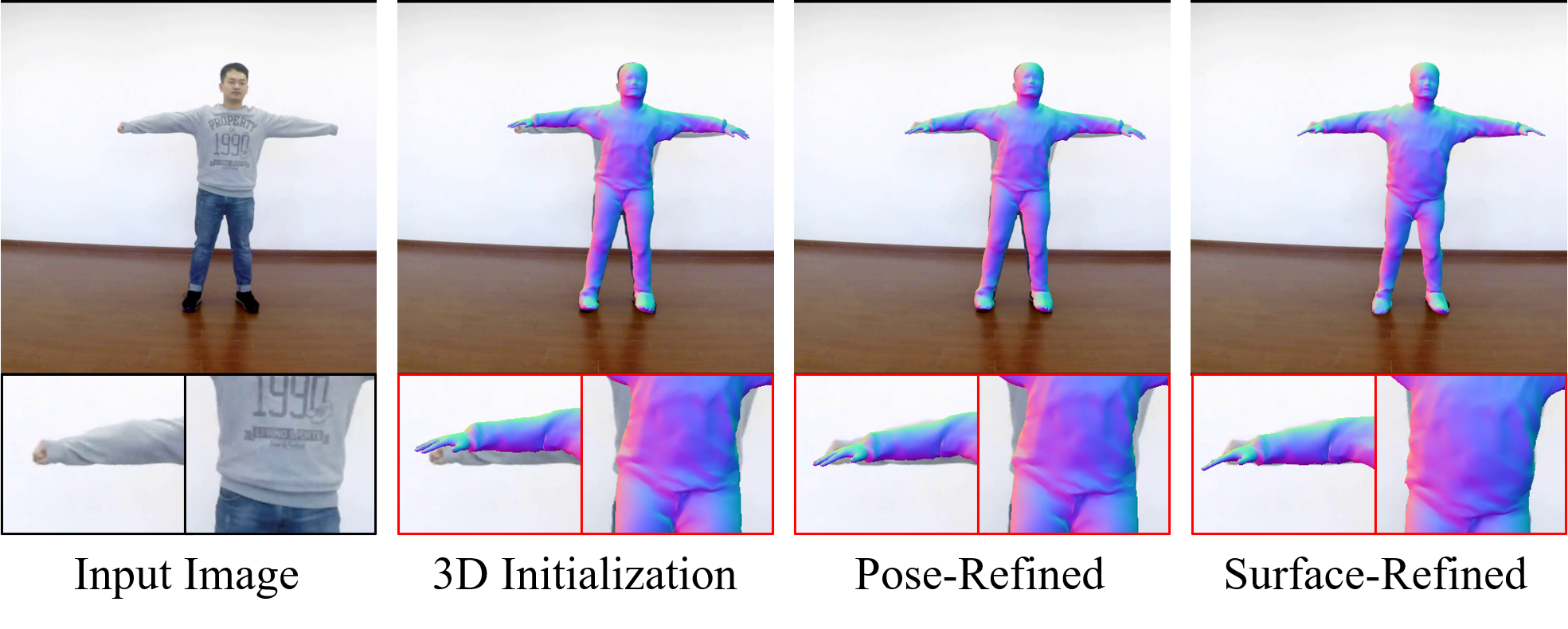}
\end{center}
   \caption{\textbf{Intermediate results after each step.} Sequential demonstration of the contribution made by each component for final performance capture results. Given a 3D initialization, we refine the poses based on 2D observations. Once the large-scale misalignment is alleviated, our surface refinement helps enforce the overlapping through non-rigid local surface deformation.}
\label{fig:ablationstep}
\end{figure}
}

\newcommand{\figureThreedpw}{
\begin{figure}
\begin{center}
\includegraphics[width=0.3\linewidth]{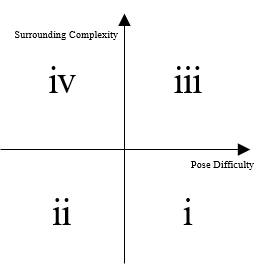}
\end{center}
   \caption{Classification of 3DPW dataset. Sequences are clustered based on their pose difficulty and complexity of surroundings through the entire video sequence.}
\label{fig:threedpw}
\end{figure}
}

\newcommand{\figureCompPIFuHDNaked}{
\begin{figure*}
\begin{center}
\includegraphics[width=\linewidth,trim=0 10 0 0, clip]{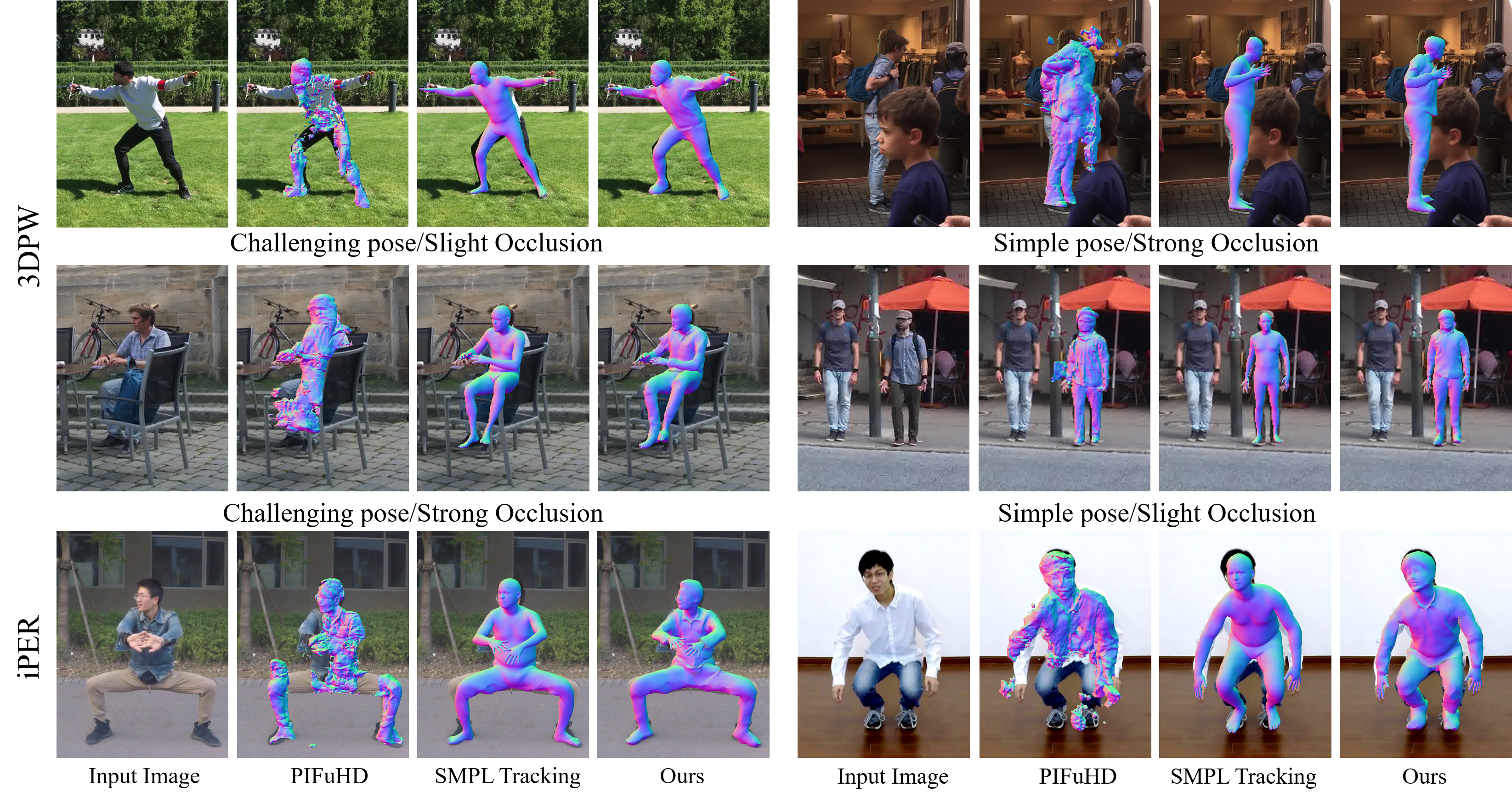}
\end{center}
\vspace{-1em}
\caption{\textbf{Qualitative comparison on 3DPW and iPER dataset.} Results of learning-based method PIFuHD, SMPL-based tracking and our method are shown. For 3DPW, we show the results in different levels of difficulties as specified in respective sub-captions. Our method produces plausible results under challenging scenarios and achieves accurate surface-image alignment. In contrast, PIFuHD's results degenerate under challenging body poses and heavy occlusions. Using SMPL model as a generic template instead of the automatically constructed template, the method fails to capture the clothing shape and deformation.}
\label{fig:comp_pifuhd_naked}
\end{figure*}
}

\newcommand{\figureCompMono}{

\begin{figure}
\begin{center}
\includegraphics[width=0.9\linewidth,trim=0 10 0 0, clip]{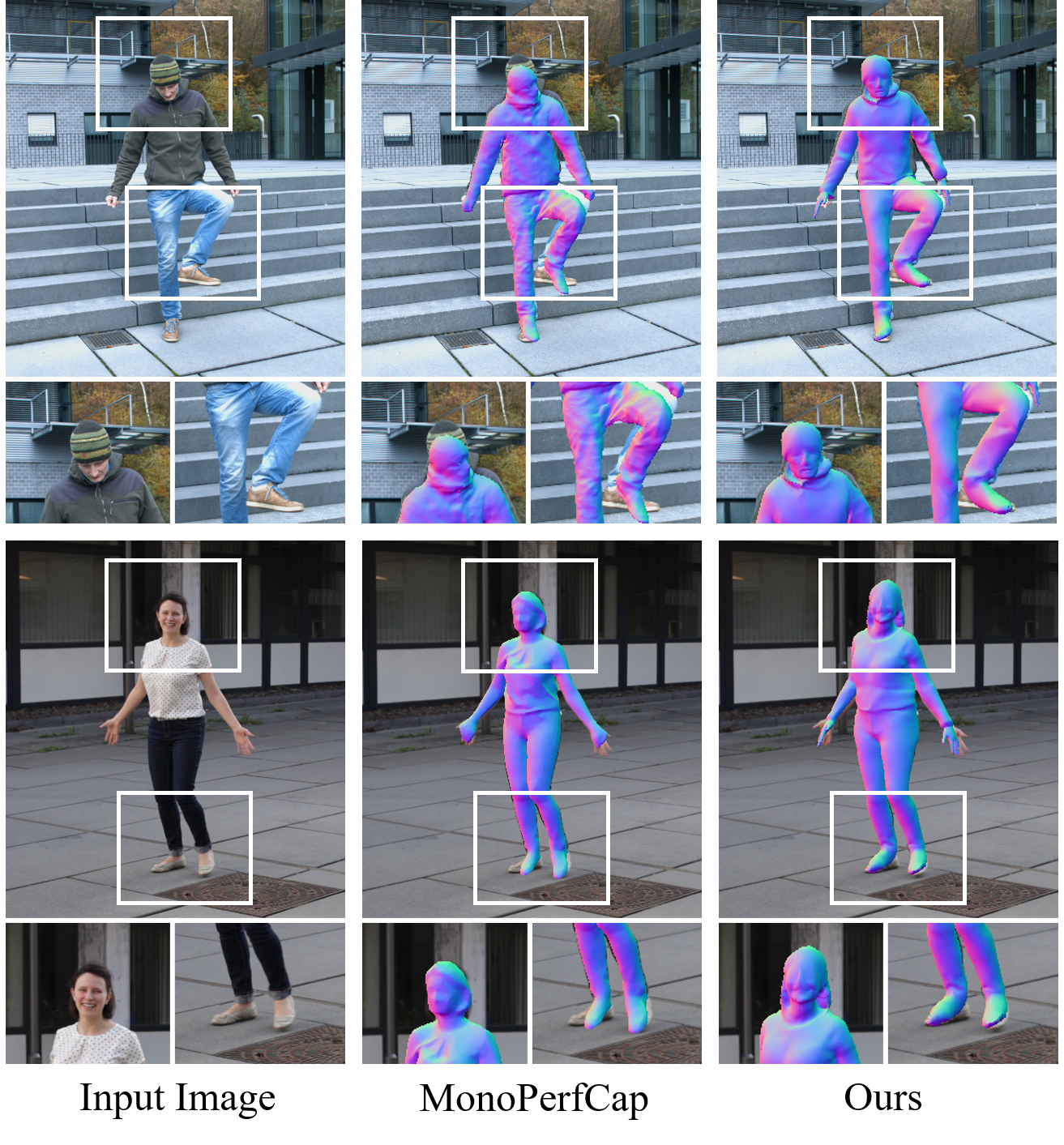}
\end{center}
\vspace{-1em}
\caption{\textbf{Qualitative comparison with MonoPerfCap.} MonoPerfCap requires a subject-specific template in addition to the monocular video, which requires multi-view capturing setup and manual efforts. In contrast, our method does not require such a template as input and achieves comparable perceptual results.}
\label{fig:comp_mono}
\end{figure}
}

\newcommand{\figureQualRes}{

\begin{figure*}
\begin{center}
\includegraphics[width=\linewidth,trim=0 0 0 0, clip]{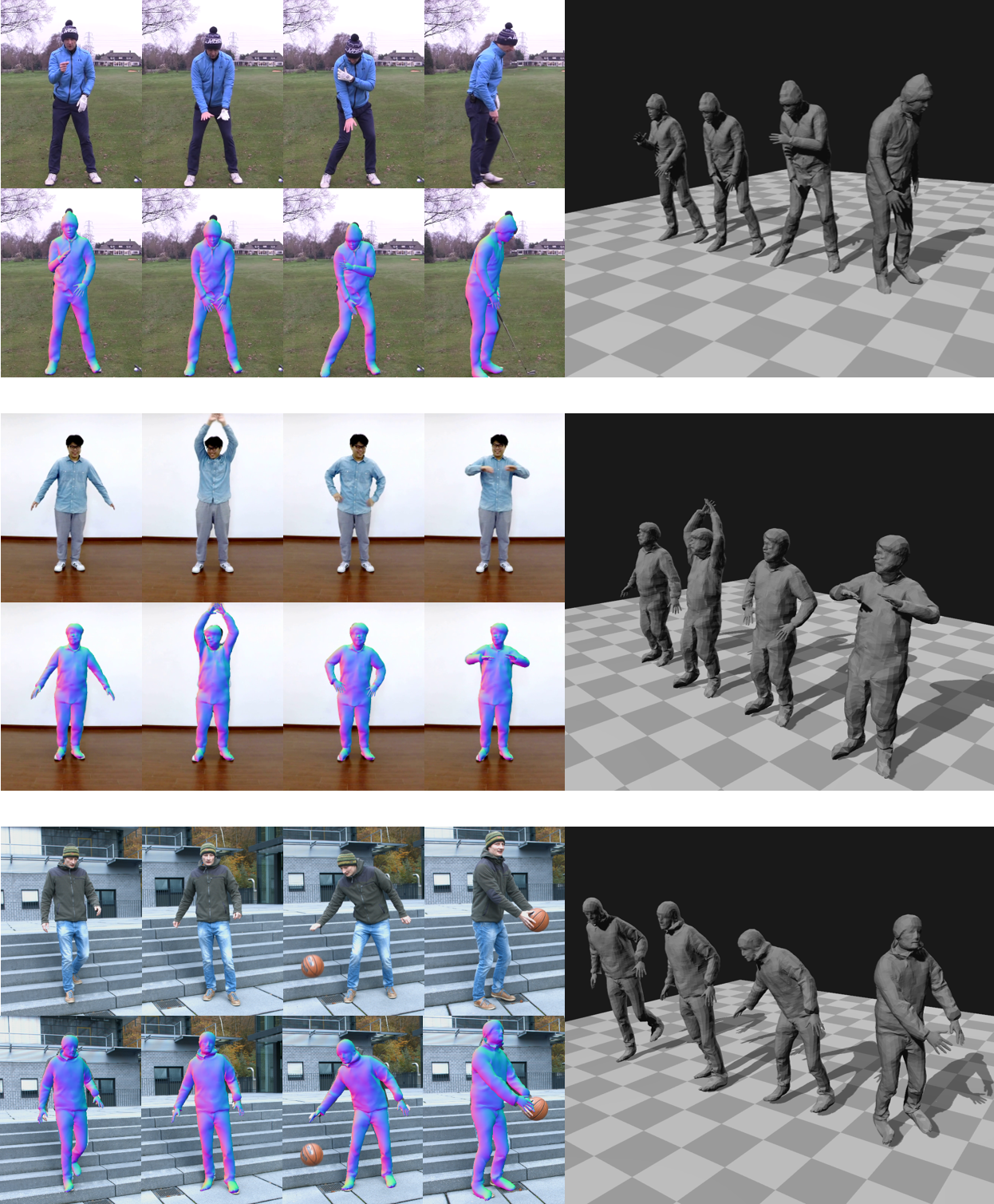}
\end{center}
   \caption{ \textbf{Additional qualitative results on iPER, MonoPerfCap and online videos.} Every two rows form a group. The top row shows the input images and the bottom row shows the estimated surfaces. On the right side, we visualize the surface from a new viewpoint.}
\label{fig:qualres}
\end{figure*}
}
\figureTeaser

In this paper, we study the problem of human performance capture from monocular in-the-wild video. It is a task of reconstructing dynamically deforming 3D shapes of human in clothing from a video featuring human motion, which is key to many applications in film/sport industry, VR/AR, and also human-computer interaction. However, reconstructing the detailed 3D geometry of human is challenging due to depth ambiguities from monocular input, the inherently complex human motions, and the high degrees of freedom in clothing deformations. 

Recently, there has been remarkable progress in this setting which can be categorized into two paradigms: learning-based approaches~\cite{saito2019pifu, saito2020pifuhd}, and tracking-based approaches~\cite{Xu:2018:MHP:3191713.3181973, habermann2019TOG, deepcap}.
Methods following the learning-based paradigm~\cite{saito2019pifu, saito2020pifuhd} learn the mapping from 2D pixels to 3D shapes using a large amount of 3D human scans. Although these methods can provide highly detailed reconstructions of human in clothing, they typically struggle in the out-of-distribution setting.
On the other hand, tracking-based methods \cite{Xu:2018:MHP:3191713.3181973, habermann2019TOG, deepcap} use a pre-rigged and subject-specific 3D template and track the template across time from a monocular video. These methods generalize better in in-the-wild setting and are more robust under challenging poses. However, acquiring the subject-specific template requires a massive multi-view capturing setup, and extensive manual efforts for post-processing, preventing such methods from being deployed to real-life applications.

In this paper, we explore to combine the best of both lines of work to build an automatic and effective system, which  can model detailed clothing deformations and is robust to in-the-wild settings without assuming a subject-specific template. Given a monocular video as input, we first leverage a learning-based single-view reconstruction method to build a 3D template from the initial frame, without additional capturing process or manual interference. We then track this template human model along time by deforming it based on 2D image observations. More specifically, we utilize a pre-trained single view human reconstruction model \cite{saito2020pifuhd} to first infer the rigid mesh model from the initial frame. We then register a generic human model SMPL \cite{SMPL:2015} onto the rigid mesh by optimizing SMPL pose and shape parameters as well as the per-vertex displacements. In this way, we obtain a parametric template model without any additional input or human effort.

Next, we track the detailed 3D geometry at each frame by estimating the deformations of the template.  The deformations are predicted by fitting the current template to 2D observations, including joints and silhouettes via a gradient-based optimization scheme. The fitting process is decomposed into two stages, similar to \cite{Xu:2018:MHP:3191713.3181973}. We first optimize the pose parameter of the model, yielding a coarse alignment to the image, and then optimize the detailed surface deformations to further refine the alignment. In this way, our method faithfully estimates both the body motion and the local surface deformations, hence produces the detailed 3D shapes at each frame.

We evaluate our proposed method on an in-the-wild human video dataset \cite{vonMarcard2018} and demonstrate that our method outperforms the state-of-the-art learning-based method especially when poses are challenging. We further compare our method with other tracking-based methods. Our method achieves on-par results but eliminates the need for multi-view capturing setup and manual efforts.

\section{Related Work}
\boldparagraph{Human Reconstruction from Multi-view/Depth} In the multi-view setting, current approaches \cite{10.1145/1360612.1360697, 580394, journals/tvcg/LiuDX10, 4178157,  10.1145/1360612.1360696, ponsmoll:hal-02162166, MuVS:3DV:2017, 10.5555/946247.946683, 1335229} estimate detailed 3D human shape based on geometric and photometric cues such as silhouette \cite{4178157}, multi-view correspondences \cite{journals/tvcg/LiuDX10}, and shading \cite{6126358}. Such methods typically require a large amount of cameras to achieve compelling results. Recent works \cite{alldieck19cvpr, alldieck20183DV, alldieck2018video} attempt to reconstruct shape from fewer cameras or pseudo multi-view setting where the subject rotates in front of a monocular camera with fixed body pose. Depth-based approaches \cite{7298631, bozic2021neuraldeformationgraphs, bozic2020deepdeform} reconstructs the human shape by fusing depth measurements across time, in order to filter sensor noise and complete occluded regions. Body prior has been introduced to handle large deformations \cite{BodyFusion,DoubleFusion,li2021posefusion, wang2020normalgan, Yu2019SimulCap, Li2020portrait}. While the aforementioned methods achieve compelling results, they require a specialized capturing setup and are hence not applicable to in-the-wild settings. In contrast, our method is capable of recovering the dynamic human shape in the wild from a monocular RGB video as the sole input.

\boldparagraph{Tracking-based Approaches with Monocular RGB} 
Tracking-based methods assume a pre-built, subject-specific 3D template model and track this model across time based on monocular video sequences \cite{habermann2019TOG, deepcap,DBLP:journals/corr/abs-2009-10711,Xu:2018:MHP:3191713.3181973, yang:hal-01344795}.
MonoPerfCap \cite{Xu:2018:MHP:3191713.3181973} captures the dynamic human with general clothing from a monocular video by fitting the template to estimated 2D and 3D human joints and 2D silhouettes.
LiveCap \cite{deepcap} further incorporates body and clothing segmentation cues to model different non-rigid deformation behaviors of skin and apparel.
DeepCap \cite{deepcap} replaces iterative optimization with deep neural networks for estimating both poses and surface deformations. 
However, obtaining the subject-specific template requires a massive multi-view capturing setup and extensive manual efforts for post-processing.
Our method achieves comparable results but does not require a pre-built template. Therefore, our method can be applied in in-the-wild settings.

\boldparagraph{Learning-based Approaches with Monocular RGB} 
Learning-based methods learn to regress 3D human shape from images by leveraging large-scale datasets. \cite{kanazawaHMR18, kolotouros2019spin, omran2018nbf, kocabas2019vibe,Guler_2019_CVPR, song2020lgd, kocabas2021pare, NIPS2017_7108, lin2021end-to-end} learn to infer body pose and shape from a single image, but only consider minimally clothed human. Various methods \cite{varol18_bodynet, Zheng2019DeepHuman, alldieck2019tex2shape, saito2020pifuhd, saito2019pifu, he2020geopifu, huang_arch_2020, zheng2020pamir, li2020monocular, e12c892272af41d7a3fce3cf5e6d1a80, Gabeur_2019_ICCV} have recently been proposed to reconstruct human in clothing. BodyNet \cite{varol18_bodynet} and DeepHuman \cite{Zheng2019DeepHuman} output human shape in the form of  occupancy voxel grids. Such representation has difficulties to capture fine details due to the high memory footprint. Neural implicit functions have been introduced to replace an explicit voxel grid and have enabled high-fidelity reconstructions from single images \cite{saito2020pifuhd, saito2019pifu, he2020geopifu, huang_arch_2020, zheng2020pamir, li2020monocular}. A major limitation of these methods is the lack of generalization to unseen poses in the wild. Our method leverages such methods to reconstruct a template of human in clothing, and generalizes well to poses beyond the training distribution by tracking the template's deformations based on image observations.

\section{Methodology}

\figurePipeline
Given a monocular video, our goal is to estimate the dynamically changing 3D surface of the subject at each frame. As shown in Fig.~\ref{fig:pipeline},
we first build a template from the initial frame of the given video, and then track how this template deforms in the successive frames based on 2D observations.

\subsection{Template Construction from Image}
At the first stage, we construct a parametric 3D template of human with clothing for the subject. The construction process only uses one frame from the input video, without requiring multi-view setup or manual efforts.

\subsubsection{Single-view Human Reconstruction} We first leverage a state-of-the-art single-view human reconstruction method \cite{saito2020pifuhd} to reconstruct the detailed shape of human from a single frame. We run the pre-trained model to obtain a rigid 3D mesh $\mathcal{S}$. The mesh surface is extracted from the implicit representation via marching cubes. 

\subsubsection{Parametric 3D Template} The resulting 3D mesh from the previous step does not yet support tracking, as vertices from independent body parts, e.g., hand and torso, might be connected. Such incorrect connectivity prevents these parts from being separated in later frames. To ensure correct connectivity of the mesh, we register a parametric human model SMPL onto the reconstructed rigid mesh $\mathcal{S}$, obtaining a parametric template. Beside guaranteeing correct connectivity, this parametric template also disentangles global skeletal deformations from local surface deformations, which enables tracking the deformations in a coarse-to-fine manner at the later stage. We first introduce the parameterization of the template, and then describe how we obtain such a template from the rigid scan.

\boldparagraph{Parameterization} SMPL \cite{SMPL:2015} is a linear statistical model of minimally clothed human bodies, mapping pose $\boldsymbol{\theta} \in \mathbb{R}^{72}$, shape $\boldsymbol{\beta} \in \mathbb{R}^{10}$ and global translation $\boldsymbol{t} \in \mathbb{R}^{3}$ to the positions of $N=6890$ vertices of a human mesh. To model details such as clothing, we further introduce the vertex displacements $\mathbf{D} \in \mathbb{R}^{N \times 3}$ as additional parameters, similar to \cite{alldieck2019learning}. The vertex positions $M$ are determined as 
\begin{align}
    M(\boldsymbol{\theta}, \boldsymbol{\beta}, \mathbf{D})=W(T(\boldsymbol{\theta}, \boldsymbol{\beta}, \mathbf{D}), \boldsymbol{\beta}, \boldsymbol{\theta}, \mathbf{W}),\\
    T(\boldsymbol{\theta}, \boldsymbol{\beta}, \mathbf{D})=\mathbf{T}_{\mu}+B_{s}(\boldsymbol{\beta})+B_{p}(\boldsymbol{\theta})+\mathbf{D},
\end{align}
where $W(\cdot)$ is the linear blend skinning algorithm which deforms a canonical mesh $T(\boldsymbol{\theta}, \boldsymbol{\beta}, \mathbf{D})$ to desired body poses based on the pre-defined skinning weights $\mathbf{W}$ and bone transformations derived from pose $\boldsymbol{\theta}$ and shape $\boldsymbol{\beta}$. The canonical mesh $T(\boldsymbol{\theta}, \boldsymbol{\beta}, \mathbf{D})$ is obtained by linearly combining shape-dependent deformations $B_{s}(\boldsymbol{\beta})$, pose-dependent deformations $B_{p}(\boldsymbol{\theta})$ and the vertex displacements $\mathbf{D}$.

\boldparagraph{Registration} To build the subject-specific template, we register the parametric template to the estimated rigid shape $\mathcal{S}$. We first use IP-Net \cite{bhatnagar2020ipnet} to obtain a minimally clothed registration, i.e. the shape $\beta$ and pose $\theta$ parameter of SMPL, as initialization. Then we optimize the vertex displacements $\boldsymbol{D}$ to minimize the following energy function:
\begin{align}
    \min_{\boldsymbol{D}}E_{\text{reg}} = E_{\text{chamfer}} + \lambda_{\text{lap}} E_{\text{lap}} + \lambda_{\text{offset}} E_{\text{offset}},
\end{align}
where $E_{\text{chamfer}}$ depicts the bi-directional Chamfer difference between the template $M$ and the reconstructed rigid mesh $\mathcal{S}$, and $E_{\text{lap}}$ and $E_{\text{offset}}$ are regularization terms weighted by $\lambda_{\text{lap}}$ and $\lambda_{\text{offset}}$ respectively. $E_{\text{lap}}$ is the Laplacian regularizer and $E_{\text{offset}}$ is the $L_{2}$ norm of the vertex displacements $\mathbf{D}$, which penalizes deviation from the minimally clothed body.

\boldparagraph{Embedded Deformation Graph} Directly optimizing vertex displacements $D$ based on 2D images is subject to errors and artifacts due to the high degrees of freedom. Thus, following \cite{sumner:2007}, we build an embedded deformation graph $\mathcal{D}$ with $K=689$ nodes, parameterized with axis angles $\mathbf{A} \in \mathbb{R}^{K\times3}$ and translations $\mathbf{T} \in \mathbb{R}^{K\times3}$. The vertex displacements are then derived from the associated deformation nodes, hence the number of parameters to be optimized is greatly reduced.

\subsection{Video-based Template Tracking}
From the previous stage, we obtain a template parameterized by the body pose $\boldsymbol{\theta}$ and shape $\boldsymbol{\beta}$, which are inherited from SMPL, and the surface deformations controlled by the deformation graph $\mathcal{D}$. To infer the human surface at successive frames, we optimize these parameters by fitting the model to image observations. This section introduces energy terms used during the optimization procedure, the initialization scheme for the parameters, and finally the optimization routine. 

\subsubsection{Energy Functions}
\boldparagraph{2D Joint Alignment $E_{\text{joint}}$} This term measures the distance between estimated 2D joints $J_{2D,\text{est}}$ from OpenPose \cite{8765346} and the 2D projection of 3D SMPL joints:
\begin{align}
    E_{\text{joint}}(\boldsymbol{\theta},\boldsymbol{\beta},\boldsymbol{t})=\sum_{i=1}^{N_{\text{joint}}} w_{i} \rho\left(\Pi\left(J(\boldsymbol{\theta}, \boldsymbol{\beta}, \boldsymbol{t} \right) )-J_{2D,\text{est}, i}\right)
\end{align}
where $J(\boldsymbol{\theta}, \boldsymbol{\beta}, \boldsymbol{t})$ is the 3D SMPL joints given the SMPL parameters. We sum up the distances for each joint $i$ overall counted joints $N_{\text{joint}}$. $\Pi$ denotes the 3D to 2D projection of joints with intrinsic camera parameters. To account for detection noise, the error terms are weighted by the corresponding detection confidence $w_{i}$. A robust Geman-McClure error function $\rho$ \cite{Geman1987StatisticalMF} is applied to down-weight outlier 2D detections. 

\boldparagraph{2D Silhouette Alignment $E_{\text{sil}}$} This term measures the overlap between the projected silhouette of the model and the estimated silhouette in the image. It serves as an important cue for inferring the surface deformations. We extract human silhouette from images using MODNet \cite{MODNet}. We calculate the overlap by comparing the difference for each pixel $p$ in the image and take an average among all pixels $\mathcal{P}$ as the final result:
\begin{align}
    &E_{\text{sil}}(\boldsymbol{\theta}, \boldsymbol{\beta}, \boldsymbol{t}, \mathcal{D}) = \frac{1}{\lvert \mathcal{P} \rvert} \sum_{p\in \mathcal{P} } \norm{S_{\text{proj},p}(\boldsymbol{\theta}, \boldsymbol{\beta}, \boldsymbol{t}, \mathcal{D}) - S_{\text{est},p}}_{2}^{2}
\end{align}
where $S_{\text{proj}}$ is the silhouette rendered by a differentiable mesh renderer \cite{ravi2020pytorch3d}.

\boldparagraph{Pose Plausibility $E_\text{prior}$} This term, proposed in \cite{bogo2016smpl}, reflects how plausible a pose is, given a pose prior learned from a large scale realistic pose corpus \cite{cmu:mocap, Loper:SIGASIA:2014}. The pose prior is modelled as a mixture of $N_\text{gauss} = 8$ Gaussian distributions with learned weights $\alpha_{j}$, mean $\boldsymbol{\mu}_{j}$, and variance $\boldsymbol{\Sigma}_{j}$. The pose plausibility is given as:
\begin{align}
E_\text{prior}(\boldsymbol{\theta}) = -\log \sum_{j=1}^{N_\text{gauss}}\alpha_{j} \mathcal{N}\left(\boldsymbol{\theta} ; \boldsymbol{\mu}_{j}, \boldsymbol{\Sigma} _{j}\right).
\end{align}

\boldparagraph{Temporal Pose Stability $E_{\text{stab}}$} This energy term is defined as the mean squared error of the current frame and the last frame 3D SMPL joints, which penalizes temporal pose jittering:
\begin{align}
E_{\text {stab}}\left( \boldsymbol{\theta},\boldsymbol{\beta},\boldsymbol{t} \right)=\sum_{i=1}^{N_{\text{joint}}}\left\|J(\boldsymbol{\theta},\boldsymbol{\beta},\boldsymbol{t})_i^f-J_i^{f-1}\right\|_{2}^{2}.
\end{align}

\boldparagraph{As-rigid-as-possible $E_{\text{arap}}$} 
This term reflects the deviation of estimated local surface deformations from rigid transformations. Here, $\mathbf{g} \in \mathbb{R}^{K\times3}$ are the original positions of the nodes in the embedded graph $\mathcal{D}$ and $\Phi(k)$ is the 1-ring neighbourhood of deformation node $k$.
\begin{align}
E_{\text{arap}}(\mathcal{D}) &= \sum_{k \in K} \sum_{l \in \Phi(k)}\left\|d_{k,l}(\mathbf{A}, \mathbf{T})\right\|_{2}^{2}, \\
d_{k, l}(\mathbf{A}, \mathbf{T})&=R\left(\mathbf{A}_{k}\right)\left(\mathbf{g}_{l}-\mathbf{g}_{k}\right)+\mathbf{T}_{k}+\mathbf{g}_{k}-\left(\mathbf{g}_{l}+\mathbf{T}_{l}\right),
\end{align}
where \noindent$R\left(\cdot\right)$ is the Rodrigues' rotation formula that computes a rotation matrix from an axis–angle representation.

\subsubsection{Initialization}

At the first frame, we estimate the global translation $\boldsymbol{t}$ based on the human bounding box in the 2D image. We initialize the rotations $\mathbf{A}$ and translations $\mathbf{T}$ of the deformation graph with zero values. 

For each frame, the global translation $\boldsymbol{t}$ and deformation graph $\mathcal{D}$ are initialized with the results from the previous frame. In terms of pose parameter $\boldsymbol{\theta}$, we obtain initial values from the state-of-the-art human mesh recovery method LGD \cite{song2020lgd}. This is crucial to recover from lost track and to prevent error accumulation during tracking. The shape parameter $\boldsymbol{\beta}$ is only optimized at the first frame and is kept fixed for the remaining. 

\subsubsection{Optimization Routine}
We decompose the optimization routine into two stages. The first stage is responsible for capturing the accurate human pose, and the second stage is designed to refine the outer surface. 

\boldparagraph{Pose Refinement.} We refine the pose $\boldsymbol{\theta}$, shape $\boldsymbol{\beta}$, and the global translation $\boldsymbol{t}$ of SMPL model to minimize the 2D joint and silhouette alignment energy terms, regularized by the pose prior and stability terms:

\begin{align}
\label{eq:poseref}
\begin{aligned}
\min_{\boldsymbol{\theta}, \boldsymbol{\beta}, \boldsymbol{t}} E_{\text{pose}}=\underbrace{E_{J_{2D}} + \lambda_{\text{sil}} E_{\text{sil}}}_{\text{data fitting}} + 
\underbrace{\lambda_{\text{stab}} E_{\text{stab}} + \lambda_{\text{prior}} E_{\text{prior}}}_{\text{regularization}}
\end{aligned}
\end{align}

\boldparagraph{Surface Refinement.} We further refine the surface deformations, represented by the deformation graph $\mathcal{D}$, to better align the parametric template to the extracted image silhouettes. This step captures the non-rigid surface deformations of apparel and skin. We optimize $\mathcal{D}$ with the silhouette alignment term and the as-rigid-as-possible regularizer, and keep other parameters fixed: 
\begin{align}
\min_{\mathcal{D}} E_{\text{surf}}=\underbrace{E_{\text{sil}}}_{\text{data fitting}}+\underbrace{\lambda_{\text{arap}} E_{\text{arap}}}_{\text {regularization}}.
\end{align}
where $\lambda_{(\cdot)}$ are the weights for the corresponding energy terms. Finally, the per-frame estimates are temporally smoothed based on a centered sliding window of 5 frames. Please refer to the supplementary material for more details.

\newcommand{\tablethreedpw}{
\begin{table}[t!]
\resizebox{\linewidth}{!}{
\begin{tabular}{|c|c|c|c|c|c|}

\hline 
\multicolumn{1}{|l|}{Pose} & Occ. & Sequence & PIFuHD & SMPL Tracking & Ours \\
\hline \hline 

\multirow{9}{*}{H} & \multirow{9}{*}{E} & 
downtown\_car & $3.57$ & $3.57$ & $\mathbf{3.26}$ \\  
& & downtown\_downstairs & $3.03$ & $2.93$ & $\mathbf{2.78}$ \\  
& & downtown\_runForBus & $4.33$ & $3.72$ & $\mathbf{3.36}$ \\  
& & downtown\_sitOnStairs & $4.41$ & $3.15$ & $\mathbf{2.98}$ \\  
 
& & downtown\_upstairs & $2.83$ & $2.74$ & $\mathbf{2.58}$ \\  
& & downtown\_walkUphill & $3.44$ & $2.83$ & $\mathbf{2.70}$ \\  
& & downtown\_weeklyMarket & $3.29$ & $3.09$ & $\mathbf{2.72}$ \\  
& & outdoors\_fencing & $6.29$ & $3.87$ & $\mathbf{3.39}$ \\  

& & Avg. & $4.09$ & $3.35$ & $\mathbf{3.08}$ \\ 
\hline \hline

\multirow{6}{*}{E} & \multirow{6}{*}{H} &
downtown\_enterShop & $2.65$ & $2.77$ & $\mathbf{2.58}$ \\ 
& & downtown\_stairs & $4.24$ & $3.79$ & $\mathbf{3.48}$ \\ 
& & downtown\_walkBridge & $3.01$ & $3.21$ & $\mathbf{2.96}$ \\ 
& & downtown\_walking & $3.38$ & $3.28$ & $\mathbf{3.15}$ \\ 

& & downtown\_windowShopping & $2.85$ & $3.17$ & $\mathbf{2.71}$ \\ 
& & Avg. & $3.21$ & $3.23$ & $\mathbf{2.98}$ \\

\hline \hline

\multirow{6}{*}{H} & \multirow{6}{*}{H} &
downtown\_bar & $4.26$ & $4.02$ & $\mathbf{3.74}$   \\  
& & downtown\_cafe & $4.00$ & $3.09$ & $\mathbf{3.03}$\\ 
& & downtown\_warmWelcome & $3.92$ & $4.05$ & $\mathbf{3.84}$ \\ 
& & flat\_actions & $6.21$ & $3.14$ & $\mathbf{3.13}$\\ 
& & office\_phoneCall & $3.34$ & $\mathbf{2.56}$ & $2.58$ \\ 

& & Avg. & $4.43$ & $3.38$ & $\mathbf{3.26}$ \\ 

\hline \hline

\multirow{5}{*}{E} & \multirow{5}{*}{E} & downtown\_arguing & $\mathbf{2.84} $ & $3.61$ & $3.39$  \\ 
& & downtown\_bus & $\mathbf{3.20}$ & $3.40$ & $\mathbf{3.20}$\\ 
& & downtown\_crossStreets & $3.02$ & $3.33$ & $\mathbf{3.00}$\\ 
& & downtown\_rampAndStairs & $\mathbf{3.01}$ & $3.16$ & $3.11$ \\ 
& & Avg. & $\mathbf{3.03}$ & $3.38$ & $3.19$ \\

\hline \hline

\multirow{1}{*}{} & \multirow{1}{*}{} &
Overall Avg. & $3.76$ & $3.34$ & $\mathbf{3.12}$ \\ 
\hline

\end{tabular}
}
\vspace{0.2em}
\caption{\textbf{Quantitative evaluation on 3DPW dataset.} Chamfer distance (cm) between reconstructed and ground-truth meshes are reported. The test dataset is divided into 4 parts based on the complexity (\textbf{E}asy and \textbf{H}ard) of pose and occlusion. Our method outperforms PIFuHD in most scenarios, especially under challenging conditions, and consistently outperforms SMPL tracking baseline.}
\label{tab:comp_threedpw}
\end{table}
}

\section{Experiments}
We compare our method with the state-of-the-art learning-based single view reconstruction method on an in-the-wild video dataset. In addition, we also conduct a qualitative evaluation with a tracking-based method that relies on a pre-scanned template mesh. Finally, we visualize the effect of individual components on the final results.

\subsection{Datasets}
We use the following datasets for evaluation and note that none of our modules is trained on any dataset below:

\figureCompPIFuHDNaked

\boldparagraph{3DPW Dataset~\cite{vonMarcard2018}} This dataset records challenging in-the-wild video sequences with accurate 3D human poses recovered by using IMUs and a moving camera. Moreover, it includes 3D scans and registered 3D people models with 18 clothing variations. By feeding the human model with the ground-truth poses and shapes, we can obtain quasi-scans to evaluate our method in terms of surface reconstruction accuracy. We evaluate our method on the test split\footnote{in case of heavy occlusions in the initial frame, we reconstruct the template from a later frame.}, and consider every 10-th frame for evaluation. Following the standard 3DPW evaluation protocol, we discard frames in which less than 6 joints are detected. In total, the evaluation set contains 24 video sequences with 3569 frames. We compute Chamfer distance (in cm) between our prediction and the ground-truth averaged over all frames for the corresponding sequence as the surface reconstruction metric. 
\tablethreedpw

\boldparagraph{MonoPerfCap Dataset \cite{Xu:2018:MHP:3191713.3181973}} This dataset contains videos of people in different garment types and actions. Subject-specific templates are also provided, which are required for tracking-based methods. In contrast, our method only uses the video not the provided template. As no per-frame ground-truth surface is provided, we resort to qualitative comparison with this baseline.

\boldparagraph{iPER Dataset \cite{liu2019liquid}} This dataset contains videos of subjects in various shapes and garments performing various actions.

\subsection{Comparison with Learning-based Method}
\label{sec:comp}
We consider PIFuHD \cite{saito2020pifuhd} as our learning-based baseline. This method is state-of-the-art in single view human reconstruction. It uses pixel-aligned features extracted from high-resolution images to guide the coarse-to-fine reconstruction. We quantitatively evaluate our method and PIFuHD on the 3DPW dataset. Tab.~\ref{tab:comp_threedpw} summarizes surface reconstruction accuracy. As can be seen, our method on average achieves approximately 17\% less error under these challenging scenarios, demonstrating the robustness of our method. This improvement is even more visible qualitatively as shown in Fig.~\ref{fig:comp_pifuhd_naked}. Our method produces plausible results even for highly dynamic poses and heavy occlusions, which are challenging for PIFuHD.

To further understand our performance, we divide the test dataset into 4 different parts with different levels of complexity in terms of pose and occlusion. Please refer to Fig.~\ref{fig:comp_pifuhd_naked} and Tab.~\ref{tab:comp_threedpw} for results in each split. In the case of \textbf{hard poses (H) but little occlusions (E)}, our approach consistently outperforms PIFuHD by a large margin. 
As for \textbf{simple poses (E) with strong occlusions (H)}, our method also shows its advantage of being able to reconstruct unseen regions. In the case where \textbf{both pose (H) and occlusion (H) are challenging}, our method is still able to produce meaningful results while PIFuHD struggles to reconstruct plausible shapes. Finally, in ideal conditions with \textbf{simple poses (E) and few occlusions (E)}, our method is less accurate than PIFuHD due to the limited resolution of the template mesh compared to PIFuHD's output.

\subsection{Comparison with Tracking-based Method}
We compare our method with MonoPerfCap \cite{Xu:2018:MHP:3191713.3181973}, a representative tracking-based method. This method captures the human performance from a monocular video, but requires a pre-built subject-specific template model. We thus conduct the evaluation on their own dataset, which provides such templates. Note that our method does not use these templates but only takes the video as input. As no ground-truth surface is provided in MonoPerfCap's dataset, we are only able to conduct a qualitative comparison. As shown in  Fig.~\ref{fig:comp_mono}, without requiring the pre-built subject-specific template, our method achieves comparable results in terms of the body pose accuracy and the fidelity of local details.

\subsection{Effect of Template Reconstruction from Image}
To verify the necessity of building the parametric 3D template for the subject, we provide an additional baseline in which we replace the reconstructed template with SMPL model as a generic template. As shown in Tab.~\ref{tab:comp_threedpw}, our method consistently outperforms this baseline (SMPL tracking). The reason is that this baseline fails to align the model to the image observations due to the shape mismatch, as displayed in Fig.~\ref{fig:comp_pifuhd_naked}. In addition, this baseline also fails to capture clothing and body details.

\figureCompMono

\subsection{User Study}
We conduct a user study to quantify the visual effects of our method.  We randomly pick 8 video clips from 3DPW and iPER dataset and ask 30 users to choose which method is preferred in terms of accuracy and perceptual fidelity. The survey in Tab.~\ref{table:user_study} indicates that our method is favored more often than baselines. PIFuHD's low performance relates to the occasional flickering when the method fails entirely.  
\begin{table}[h]
\centering
\setlength\tabcolsep{3.3pt}
\begin{tabular}{lccc}

\hline
& PIFuHD & SMPL Tracking & Ours   \\ \hline

Vote rate &5.83\%  &8.33\%  &85.83\%   \\
\hline
\end{tabular}
\vspace{0.5em}
\caption{\textbf{User study.} Vote rate in average.}
\label{table:user_study}
\end{table}

\subsection{Effect of Optimization Stages}
We now illustrate the effect of main steps during tracking, namely, 3D pose initialization, pose and surface refinement. First, we use the pose from the previous frame to replace the learned 3D initialization. As shown in Fig.~\ref{fig:ablationfinal}, while the estimated surface still aligns with the 2D joints and silhouette, the 3D pose is implausible, demonstrating that the learned 3D pose initialization is important to tackle the inherent depth ambiguity from a single view. Secondly, we keep all components but skip the pose refinement stage. This leads to notable misalignment to the image observation, e.g., the hands in Fig.~\ref{fig:ablationfinal}. Finally, removing the surface refinement step from the full pipeline leads to even more notable misalignment, e.g., at the boundary of the pants. The complete pipeline achieves accurate surface-to-image alignment without suffering from degenerated poses. 

\figureAblationFinal

\subsection{Qualitative Results}
We show qualitative results on different datasets in Fig.~\ref{fig:qualres} with overlaid images and the ones from 3D free-view points. Our approach can generalize to online videos with different garments, contexts and gestures. Please refer to the supplementary materials and video for more samples. 

\figureQualRes

\section{Conclusion}
We propose a method to estimate 3D human shape in clothing from a sole monocular video. Compared to tracking-based methods, our method does not require a pre-scanned template thus can be applied more broadly, such as internet videos. Compared to learning-based ones, our method generalizes better to in-the-wild videos with natural and dynamic poses. Our attempt demonstrates the potential of 
integrating tracking and learning-based methods to tackle the problem of 3D human reconstruction.

{\small
\noindent\textbf{Acknowledgements:} Xu Chen was supported by the Max Planck ETH Center for Learning Systems. 
}

{\small
\bibliographystyle{ieee}
\bibliography{bibliography_long,egbib}

\begin{thebibliography}{10}\itemsep=-1pt

\bibitem{cmu:mocap}
\url{http://mocap.cs.cmu.edu}.

\bibitem{alldieck19cvpr}
T.~Alldieck, M.~Magnor, B.~L. Bhatnagar, C.~Theobalt, and G.~Pons-Moll.
\newblock Learning to reconstruct people in clothing from a single {RGB}
  camera.
\newblock In {\em Proc. IEEE Conf. on Computer Vision and Pattern Recognition
  (CVPR)}, 2019.

\bibitem{alldieck2019learning}
T.~Alldieck, M.~Magnor, B.~L. Bhatnagar, C.~Theobalt, and G.~Pons-Moll.
\newblock Learning to reconstruct people in clothing from a single {RGB}
  camera.
\newblock In {\em Proc. IEEE Conf. on Computer Vision and Pattern Recognition
  (CVPR)}, 2019.

\bibitem{alldieck20183DV}
T.~Alldieck, M.~Magnor, W.~Xu, C.~Theobalt, and G.~Pons-Moll.
\newblock Detailed human avatars from monocular video.
\newblock In {\em International Conference on 3D Vision (3DV)}, 2018.

\bibitem{alldieck2018video}
T.~Alldieck, M.~Magnor, W.~Xu, C.~Theobalt, and G.~Pons-Moll.
\newblock Video based reconstruction of 3d people models.
\newblock In {\em Proc. IEEE Conf. on Computer Vision and Pattern Recognition
  (CVPR)}, 2018.

\bibitem{alldieck2019tex2shape}
T.~Alldieck, G.~Pons-Moll, C.~Theobalt, and M.~Magnor.
\newblock Tex2shape: Detailed full human body geometry from a single image.
\newblock In {\em Proc. of the IEEE International Conf. on Computer Vision
  (ICCV)}, 2019.

\bibitem{bhatnagar2020ipnet}
B.~L. Bhatnagar, C.~Sminchisescu, C.~Theobalt, and G.~Pons-Moll.
\newblock Combining implicit function learning and parametric models for 3d
  human reconstruction.
\newblock In {\em Proc. of the European Conf. on Computer Vision (ECCV)}, 2020.

\bibitem{bogo2016smpl}
F.~Bogo, A.~Kanazawa, C.~Lassner, P.~Gehler, J.~Romero, and M.~J. Black.
\newblock Keep it {SMPL}: Automatic estimation of {3D} human pose and shape
  from a single image.
\newblock In {\em Proc. of the European Conf. on Computer Vision (ECCV)}, 2016.

\bibitem{bozic2021neuraldeformationgraphs}
A.~Bo{\v{z}}i{\v{c}}, P.~Palafox, M.~Zollh{\"o}fer, J.~Thies, A.~Dai, and
  M.~Nie{\ss}ner.
\newblock Neural deformation graphs for globally-consistent non-rigid
  reconstruction.
\newblock In {\em Proc. IEEE Conf. on Computer Vision and Pattern Recognition
  (CVPR)}, 2021.

\bibitem{bozic2020deepdeform}
A.~Bo{\v{z}}i{\v{c}}, M.~Zollh{\"o}fer, C.~Theobalt, and M.~Nie{\ss}ner.
\newblock Deepdeform: Learning non-rigid rgb-d reconstruction with
  semi-supervised data.
\newblock In {\em Proc. IEEE Conf. on Computer Vision and Pattern Recognition
  (CVPR)}, 2020.

\bibitem{8765346}
Z.~{Cao}, G.~{Hidalgo Martinez}, T.~{Simon}, S.~{Wei}, and Y.~A. {Sheikh}.
\newblock Openpose: Realtime multi-person 2d pose estimation using part
  affinity fields.
\newblock {\em IEEE Transactions on Pattern Analysis and Machine Intelligence},
  2019.

\bibitem{10.1145/1360612.1360697}
E.~de~Aguiar, C.~Stoll, C.~Theobalt, N.~Ahmed, H.-P. Seidel, and S.~Thrun.
\newblock Performance capture from sparse multi-view video.
\newblock {\em ACM Trans. on Graphics}, 27(3):1–10, 2008.

\bibitem{Gabeur_2019_ICCV}
V.~Gabeur, J.-S. Franco, X.~Martin, C.~Schmid, and G.~Rogez.
\newblock Moulding humans: Non-parametric 3d human shape estimation from single
  images.
\newblock In {\em Proc. of the IEEE International Conf. on Computer Vision
  (ICCV)}, 2019.

\bibitem{Geman1987StatisticalMF}
S.~Geman and D.~E. McClure.
\newblock Statistical methods for tomographic image reconstruction.
\newblock 1987.

\bibitem{Guler_2019_CVPR}
R.~A. Guler and I.~Kokkinos.
\newblock Holopose: Holistic 3d human reconstruction in-the-wild.
\newblock In {\em Proc. IEEE Conf. on Computer Vision and Pattern Recognition
  (CVPR)}, 2019.

\bibitem{habermann2019TOG}
M.~Habermann, W.~Xu, , M.~Zollhoefer, G.~Pons-Moll, and C.~Theobalt.
\newblock Livecap: Real-time human performance capture from monocular video.
\newblock {\em ACM Trans. on Graphics}, 2019.

\bibitem{deepcap}
M.~Habermann, W.~Xu, M.~Zollhoefer, G.~Pons-Moll, and C.~Theobalt.
\newblock Deepcap: Monocular human performance capture using weak supervision.
\newblock In {\em Proc. IEEE Conf. on Computer Vision and Pattern Recognition
  (CVPR)}, 2020.

\bibitem{he2020geopifu}
T.~He, J.~Collomosse, H.~Jin, and S.~Soatto.
\newblock Geo-pifu: Geometry and pixel aligned implicit functions for
  single-view human reconstruction.
\newblock {\em arXiv.org}, 2006.08072, 2020.

\bibitem{1335229}
A.~Hilton and J.~Starck.
\newblock Multiple view reconstruction of people.
\newblock In {\em Proceedings. 2nd International Symposium on 3D Data
  Processing, Visualization and Transmission, 2004. 3DPVT 2004.}, pages
  357--364, 2004.

\bibitem{MuVS:3DV:2017}
Y.~Huang, F.~Bogo, C.~Lassner, A.~Kanazawa, P.~V. Gehler, J.~Romero, I.~Akhter,
  and M.~J. Black.
\newblock Towards accurate marker-less human shape and pose estimation over
  time.
\newblock In {\em International Conference on 3D Vision (3DV)}, pages 421--430,
  2017.

\bibitem{huang_arch_2020}
Z.~Huang, Y.~Xu, C.~Lassner, H.~Li, and T.~Tung.
\newblock {ARCH}: {Animatable} {Reconstruction} of {Clothed} {Humans}.
\newblock In {\em Proc. IEEE Conf. on Computer Vision and Pattern Recognition
  (CVPR)}, Seattle, Washington, 2020. IEEE.

\bibitem{580394}
T.~Kanade, P.~Rander, and P.~Narayanan.
\newblock Virtualized reality: constructing virtual worlds from real scenes.
\newblock {\em IEEE MultiMedia}, 4(1):34--47, 1997.

\bibitem{kanazawaHMR18}
A.~Kanazawa, M.~J. Black, D.~W. Jacobs, and J.~Malik.
\newblock End-to-end recovery of human shape and pose.
\newblock In {\em Proc. IEEE Conf. on Computer Vision and Pattern Recognition
  (CVPR)}, 2018.

\bibitem{MODNet}
Z.~Ke, K.~Li, Y.~Zhou, Q.~Wu, X.~Mao, Q.~Yan, and R.~W. Lau.
\newblock Is a green screen really necessary for real-time portrait matting?
\newblock {\em arXiv.org}, 2011.11961, 2020.

\bibitem{kocabas2019vibe}
M.~Kocabas, N.~Athanasiou, and M.~J. Black.
\newblock Vibe: Video inference for human body pose and shape estimation.
\newblock In {\em Proc. IEEE Conf. on Computer Vision and Pattern Recognition
  (CVPR)}, 2020.

\bibitem{kocabas2021pare}
M.~Kocabas, C.-H.~P. Huang, O.~Hilliges, and M.~J. Black.
\newblock Pare: Part attention regressor for 3d human body estimation.
\newblock {\em arXiv.org}, 2104.08527, 2021.

\bibitem{kolotouros2019spin}
N.~Kolotouros, G.~Pavlakos, M.~J. Black, and K.~Daniilidis.
\newblock Learning to reconstruct 3d human pose and shape via model-fitting in
  the loop.
\newblock In {\em Proc. of the IEEE International Conf. on Computer Vision
  (ICCV)}, 2019.

\bibitem{li2020monocular}
R.~Li, Y.~Xiu, S.~Saito, Z.~Huang, K.~Olszewski, and H.~Li.
\newblock Monocular real-time volumetric performance capture.
\newblock In {\em Proc. of the European Conf. on Computer Vision (ECCV)}, pages
  49--67. Springer, 2020.

\bibitem{Li2020portrait}
Z.~Li, T.~Yu, C.~Pan, Z.~Zheng, and Y.~Liu.
\newblock Robust 3d self-portraits in seconds.
\newblock In {\em Proc. IEEE Conf. on Computer Vision and Pattern Recognition
  (CVPR)}, 2020.

\bibitem{li2021posefusion}
Z.~Li, T.~Yu, Z.~Zheng, K.~Guo, and Y.~Liu.
\newblock Posefusion: Pose-guided selective fusion for single-view human
  volumetric capture.
\newblock In {\em Proc. IEEE Conf. on Computer Vision and Pattern Recognition
  (CVPR)}, 2021.

\bibitem{lin2021end-to-end}
K.~Lin, L.~Wang, and Z.~Liu.
\newblock End-to-end human pose and mesh reconstruction with transformers.
\newblock In {\em Proc. IEEE Conf. on Computer Vision and Pattern Recognition
  (CVPR)}, 2021.

\bibitem{liu2019liquid}
W.~Liu, Z.~Piao, J.~Min, W.~Luo, L.~Ma, and S.~Gao.
\newblock Liquid warping gan: A unified framework for human motion imitation,
  appearance transfer and novel view synthesis.
\newblock {\em arXiv.org}, 1909.12224, 2019.

\bibitem{journals/tvcg/LiuDX10}
Y.~Liu, Q.~Dai, and W.~Xu.
\newblock A point-cloud-based multiview stereo algorithm for free-viewpoint
  video.
\newblock {\em IEEE Trans. Vis. Comput. Graph.}, 16(3):407--418, 2010.

\bibitem{SMPL:2015}
M.~Loper, N.~Mahmood, J.~Romero, G.~Pons-Moll, and M.~J. Black.
\newblock {SMPL}: A skinned multi-person linear model.
\newblock {\em ACM Trans. on Graphics}, 34(6):248:1--248:16, Oct. 2015.

\bibitem{Loper:SIGASIA:2014}
M.~M. Loper, N.~Mahmood, and M.~J. Black.
\newblock {MoSh}: Motion and shape capture from sparse markers.
\newblock {\em ACM Transactions on Graphics, (Proc. SIGGRAPH Asia)},
  33(6):220:1--220:13, Nov. 2014.

\bibitem{e12c892272af41d7a3fce3cf5e6d1a80}
R.~Natsume, S.~Saito, Z.~Huang, W.~Chen, C.~Ma, H.~Li, and S.~Morishima.
\newblock Siclope: Silhouette-based clothed people.
\newblock In {\em Proc. IEEE Conf. on Computer Vision and Pattern Recognition
  (CVPR)}, 2019.

\bibitem{7298631}
R.~A. Newcombe, D.~Fox, and S.~M. Seitz.
\newblock Dynamicfusion: Reconstruction and tracking of non-rigid scenes in
  real-time.
\newblock In {\em Proc. IEEE Conf. on Computer Vision and Pattern Recognition
  (CVPR)}, pages 343--352, 2015.

\bibitem{omran2018nbf}
M.~Omran, C.~Lassner, G.~Pons-Moll, P.~V. Gehler, and B.~Schiele.
\newblock Neural body fitting: Unifying deep learning and model-based human
  pose and shape estimation.
\newblock In {\em International Conference on 3D Vision (3DV)}, 2018.

\bibitem{ponsmoll:hal-02162166}
G.~Pons-Moll, S.~Pujades, S.~Hu, and M.~J. Black.
\newblock {ClothCap: Seamless 4D Clothing Capture and Retargeting}.
\newblock {\em ACM Trans. on Graphics}, 36(4):1--15, 2017.

\bibitem{ravi2020pytorch3d}
N.~Ravi, J.~Reizenstein, D.~Novotny, T.~Gordon, W.-Y. Lo, J.~Johnson, and
  G.~Gkioxari.
\newblock Accelerating 3d deep learning with pytorch3d.
\newblock {\em arXiv.org}, 2007.08501, 2020.

\bibitem{saito2019pifu}
S.~Saito, Z.~Huang, R.~Natsume, S.~Morishima, A.~Kanazawa, and H.~Li.
\newblock Pifu: Pixel-aligned implicit function for high-resolution clothed
  human digitization.
\newblock In {\em Proc. of the IEEE International Conf. on Computer Vision
  (ICCV)}, 2019.

\bibitem{saito2020pifuhd}
S.~Saito, T.~Simon, J.~Saragih, and H.~Joo.
\newblock Pifuhd: Multi-level pixel-aligned implicit function for
  high-resolution 3d human digitization.
\newblock In {\em Proc. IEEE Conf. on Computer Vision and Pattern Recognition
  (CVPR)}, 2020.

\bibitem{song2020lgd}
J.~Song, X.~Chen, and O.~Hilliges.
\newblock Human body model fitting by learned gradient descent.
\newblock In {\em Proc. of the European Conf. on Computer Vision (ECCV)}, 2020.

\bibitem{10.5555/946247.946683}
J.~Starck and A.~Hilton.
\newblock Model-based multiple view reconstruction of people.
\newblock In {\em Proc. of the IEEE International Conf. on Computer Vision
  (ICCV)}, USA, 2003. IEEE Computer Society.

\bibitem{4178157}
J.~Starck and A.~Hilton.
\newblock Surface capture for performance-based animation.
\newblock {\em IEEE Computer Graphics and Applications}, 27(3):21--31, 2007.

\bibitem{sumner:2007}
R.~Sumner, J.~Schmid, and M.~Pauly.
\newblock Embedded deformation for shape manipulation.
\newblock {\em ACM Trans. on Graphics}, 26, 07 2007.

\bibitem{NIPS2017_7108}
H.-Y. Tung, H.-W. Tung, E.~Yumer, and K.~Fragkiadaki.
\newblock Self-supervised learning of motion capture.
\newblock In I.~Guyon, U.~V. Luxburg, S.~Bengio, H.~Wallach, R.~Fergus,
  S.~Vishwanathan, and R.~Garnett, editors, {\em Advances in Neural Information
  Processing Systems (NeurIPS)}, pages 5236--5246. Curran Associates, Inc.,
  2017.

\bibitem{varol18_bodynet}
G.~Varol, D.~Ceylan, B.~Russell, J.~Yang, E.~Yumer, I.~Laptev, and C.~Schmid.
\newblock {BodyNet}: Volumetric inference of {3D} human body shapes.
\newblock In {\em Proc. of the European Conf. on Computer Vision (ECCV)}, 2018.

\bibitem{10.1145/1360612.1360696}
D.~Vlasic, I.~Baran, W.~Matusik, and J.~Popovi\'{c}.
\newblock Articulated mesh animation from multi-view silhouettes.
\newblock {\em ACM Trans. on Graphics}, 27(3):1–9, 2008.

\bibitem{vonMarcard2018}
T.~von Marcard, R.~Henschel, M.~Black, B.~Rosenhahn, and G.~Pons-Moll.
\newblock Recovering accurate 3d human pose in the wild using imus and a moving
  camera.
\newblock In {\em Proc. of the European Conf. on Computer Vision (ECCV)}, 2018.

\bibitem{wang2020normalgan}
L.~Wang, X.~Zhao, T.~Yu, S.~Wang, and Y.~Liu.
\newblock Normalgan: Learning detailed 3d human from a single rgb-d image.
\newblock In {\em Proc. of the European Conf. on Computer Vision (ECCV)}, 2020.

\bibitem{6126358}
C.~Wu, K.~Varanasi, Y.~Liu, H.-P. Seidel, and C.~Theobalt.
\newblock Shading-based dynamic shape refinement from multi-view video under
  general illumination.
\newblock In {\em Proc. of the IEEE International Conf. on Computer Vision
  (ICCV)}, pages 1108--1115, 2011.

\bibitem{DBLP:journals/corr/abs-2009-10711}
D.~Xiang, F.~Prada, C.~Wu, and J.~K. Hodgins.
\newblock Monoclothcap: Towards temporally coherent clothing capture from
  monocular {RGB} video.
\newblock {\em arXiv.org}, 2009.10711, 2020.

\bibitem{Xu:2018:MHP:3191713.3181973}
W.~Xu, A.~Chatterjee, M.~Zollh\"{o}fer, H.~Rhodin, D.~Mehta, H.-P. Seidel, and
  C.~Theobalt.
\newblock Monoperfcap: Human performance capture from monocular video.
\newblock {\em ACM Trans. on Graphics}, 37(2):27:1--27:15, May 2018.

\bibitem{yang:hal-01344795}
J.~Yang, J.-S. Franco, F.~H{\'e}troy-Wheeler, and S.~Wuhrer.
\newblock {Estimation of Human Body Shape in Motion with Wide Clothing}.
\newblock In {\em {ECCV 2016 - European Conference on Computer Vision 2016}},
  2016.

\bibitem{BodyFusion}
T.~Yu, K.~Guo, F.~Xu, Y.~Dong, Z.~Su, J.~Zhao, J.~Li, Q.~Dai, and Y.~Liu.
\newblock Bodyfusion: Real-time capture of human motion and surface geometry
  using a single depth camera.
\newblock In {\em Proc. of the IEEE International Conf. on Computer Vision
  (ICCV)}, 2017.

\bibitem{DoubleFusion}
T.~Yu, Z.~Zheng, K.~Guo, J.~Zhao, Q.~Dai, H.~Li, G.~Pons-Moll, and Y.~Liu.
\newblock Doublefusion: Real-time capture of human performances with inner body
  shapes from a single depth sensor.
\newblock In {\em Proc. IEEE Conf. on Computer Vision and Pattern Recognition
  (CVPR)}, 2018.

\bibitem{Yu2019SimulCap}
T.~Yu, Z.~Zheng, Y.~Zhong, J.~Zhao, Q.~Dai, G.~Pons-Moll, and Y.~Liu.
\newblock Simulcap : Single-view human performance capture with cloth
  simulation.
\newblock In {\em Proc. IEEE Conf. on Computer Vision and Pattern Recognition
  (CVPR)}, 2019.

\bibitem{zheng2020pamir}
Z.~Zheng, T.~Yu, Y.~Liu, and Q.~Dai.
\newblock Pamir: Parametric model-conditioned implicit representation for
  image-based human reconstruction.
\newblock {\em arXiv.org}, 2007.03858, 2020.

\bibitem{Zheng2019DeepHuman}
Z.~Zheng, T.~Yu, Y.~Wei, Q.~Dai, and Y.~Liu.
\newblock Deephuman: 3d human reconstruction from a single image.
\newblock In {\em Proc. of the IEEE International Conf. on Computer Vision
  (ICCV)}, 2019.

\end{thebibliography}
}

\end{document}